# Development of an EKF-based Localization Algorithm Using Compass Sensor and LRF

T. T. Hoang, P. M. Duong, N. T. T. Van, D. A. Viet and T. Q. Vinh
VNU University of Engineering and Technology
Hanoi, Vietnam
thuanhoang@donga.edu.vn

*Abstract*— **This paper presents the implementation of a perceptual system for a mobile robot. The system is designed and installed with modern sensors and multi-point communication channels. The goal is to equip the robot with a high level of perception to support a wide range of navigating applications including Internet-based telecontrol, semi-autonomy, and autonomy. Due to uncertainties of acquiring data, a sensor fusion model is developed, in which heterogeneous measured data including odometry, compass heading and laser range is combined to get an optimal estimate in a statistical sense. The combination is carried out by an extended Kalman filter. Experimental results indicate that based on the system, the robot localization is enhanced and sufficient for navigation tasks.**

*Index Terms*— **sensor fusion; Kalman Filter; localization;**

## I. Introduction

Perceptual system is the key to the intelligence of a mobile robot. When a mobile robot travels in an unknown or partially known environment, it must understand the environment for obstacle avoidance or path planning. The perception of robot is done by taking measurements using various sensors and then extracting meaningful information from those measurements.

Based on advanced material technologies, modern sensors can be nowadays equipped for the mobile robot such as optical incremental encoders, heading sensors, ultrasonic sensors, infrared sensors, laser range finders and vision systems. These sensors are selected so as to accord with the goal of application, the specific constraints of the working environment, and the individual properties of the sensors themselves. Nevertheless, there is no single sensor which can adequately capture all relevant features of a real environment. It is necessary to combine the data from different sensors into a process known as *sensor fusion*. The expectation is that the fused data is more informative and synthetic than the original.

Several methods have been reported to cope with this trend. Durrant-Whyte has developed a multi-Bayesian estimation technique for combining touch and stereo sensing [1], [2]. Tang and Lee proposed a generic framework that employed a sensor-independent, feature-based relational model to represent information acquired by various sensors [3]. In [4], a Kalman filter update equation was developed to obtain the correspondence of a line segment to a model, and this correspondence was then used to correct position estimation. In [5], an extended Kalman filter was conducted to manipulate image and spatial uncertainties.

In this work, we develop a multi-sensor perceptual system for the mobile robot. Sensors include but not limit to optical quadrature encoders, compass sensors, ultrasonic sensors, laser range finders, global positioning systems (GPS) and vision systems. The goal is to equip the robot with diverse levels of perception to support a wide range of navigating applications including Internet-based telecontrol, semi-autonomy, and autonomy. At this stage of research, we use the optical quadrature encoders for position measurement, compass sensor for deflect angle calculation and laser range finder (LRF) for object-boundary detection. These types of data have their strengths and are often used in robotic applications, but combining them provides even more useful information. This combined approach deserves further investigation. In our system, an extended Kalman filter (EKF) has been designed. It fuses the raw measurement data of optical quadrature encoders, the compass sensor, and the LRF to obtain optimal estimation of robot positions while reduces uncertainties in measurements. The proposed fusing algorithm also provides constraints that filter out unreliable data from the sensor readings. Outputs of EKF combined with the boundaries of objects detected from LRF ensure the success navigation of the mobile robot in indoor environments.

The paper is arranged as follows. Details of the perceptual system are described in Section II. The algorithm for sensor fusion using EKF is explained in Section III. Section IV introduces experimental results. The paper concludes with an evaluation of the system, with respect to its strengths and weaknesses, and with suggestions of possible future developments.

## II. Sensor System Design

The sensor system consists of an Omni-directional camera, a GPS, a compass sensor, a LRF, three quadrature encoders and eight ultrasonic sensors. Fig.1 describes the sensors in relation with actuators and communication channels in a mobile robot. The communication is performed via low-rate and high-rate channels. The low-rate channel with standards of RS-485 is developed by an on-board 60MHz Microchip dsPIC30F4011-based micro-controller and MODBUS protocol for multi-point interface. The high-rate channels use the USB-to-COM and IEEE-1394/firewire available commercial ports.

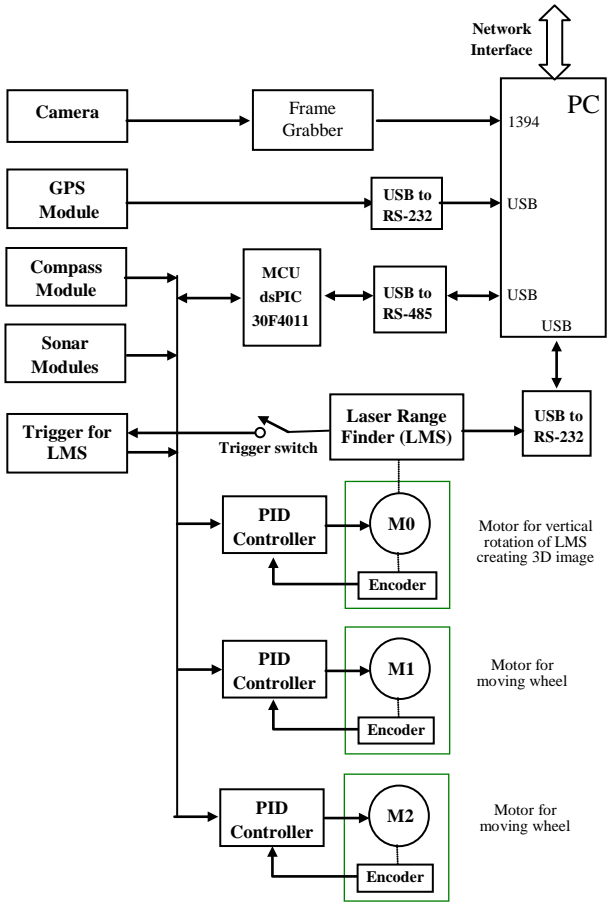

Figure 1. Sensors in relation with actuators and communication channels in our mobile robot

In the system, *optical quadrature encoders* are used. An optical encoder is basically a mechanical light chopper that produces a certain number of sine or square wave pulses for each shaft revolution. As the diameter of wheel and the gear coefficient are known, the angular position and speed of wheel can be calculated. In the quadrature encoder, a second light chopper is placed 90 degrees shifted with respect to the original resulting in twin square waves as shown in fig.2. Observed phase relationship between waves is employed to determine the direction of the rotation. In the system, measurements from encoders are used as input data for positioning and feedback for a closed-loop motor speed controller.

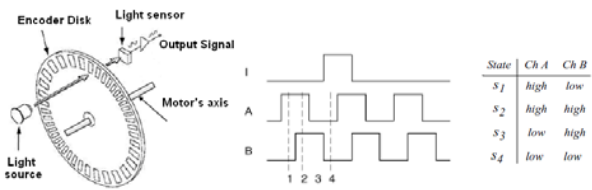

Figure 2. Optical encoder structure and output pulses

The heading sensor is used to determine the robot orientation. This sensory module contains a CMPS03 compass sensor operating based on Hall effect with heading resolution of 0.1° (fig.3). The module has two axes, x and y. Each axis reports the strength of the magnetic field component paralleled to it. The microcontroller connected to the module uses synchronous serial communication to get axis measurements [6].

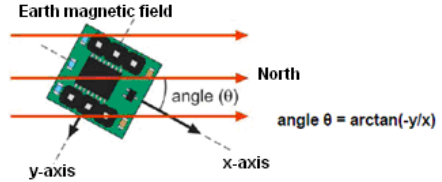

Figure 3. Heading module and output data

The GPS is mainly applied for positioning in the outdoor environment. A HOLUX GPS UB-93 module is used [7]. Due to the presence of networking in our system, an Assisted GPS (A-GPS) can be also used in order to locate and utilize satellite information of the network in the poor signal condition.

The system provides eight SFR-05 ultrasonic sensors split into four arrays, two on each, arranged at four sides of the robot. The measuring range is from 0.04m to 4m.

The vision system is detachable and mounted on the top of the robot. It mainly consists of an Omni-directional digital camera Hyper-Omni Vision SOIOS 55 with a high-rate IEEE-1394 data transfer line. With a 360-degree field of view, the Omni-directional camera is a good vision sensor for landmark-based navigation [8].

Last but not least, a 3D laser range finder (LRF) with a range from 0.04m to 80m is developed for the system. Its operation is based on the time-of-flight measurement principle. A single laser pulse is emitted out and reflected by an object surface within the range of the sensor. The lapsed time between emission and reception of the laser pulse is used to calculate the distance between the object and the sensor. By an integrated rotating mirror, the laser pulses sweep a radial range in its front so that a 2D field/area is defined (fig.4).

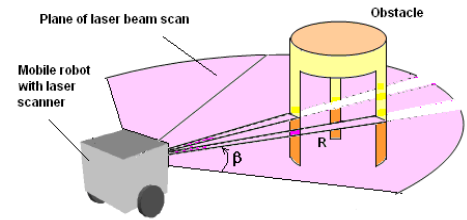

Figure 4. Two-dimension laser scanning plane [10].

Due to the fixation of the pitching angle in the scanning plane, the information of 2D image may be, in certain cases, insufficient for obstacle detection (fig.4). In those situations, a 3D image is necessary. As the 2D scanner is popular and low-cost, building a 3D LRF from the 2D is usually a prior choice [9][10]. In our system, a 3D LRF is developed based on the 2D SICK-LMS 221 [9]. It has the view angle of 180° in horizontal and 25° in vertical. During the measurement time, a set of deflect angle $\beta$ and distance $\rho$ values are received. The set ($\beta$, $\rho$) is then combined with the pitching angle $\alpha$ to create the 3D data. Based on these data, we can define the Cartesian coordinates of one point in the space.

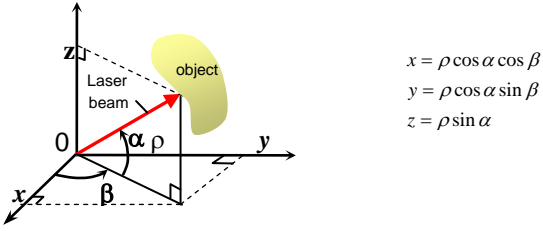

Figure 5. Determine the coordinates of one point in 3D space [10].

Fig.6 shows the proposed sensor system implemented in a mobile robot developed by our laboratory.

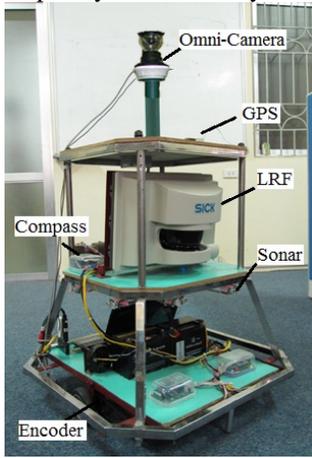

Figure 6. The implemented sensor system

## III. SENSOR FUSION

The proposed sensor platform equips the robot with the ability to perceive many parameters of the environment. Their combination, however, presents even more useful information. In this work, the raw data of three different types of sensors including the compass sensor, the quadrature encoder and the LRF is syndicated inside an EKF. The aim is to determine the robot position during operation as accurately as possible.

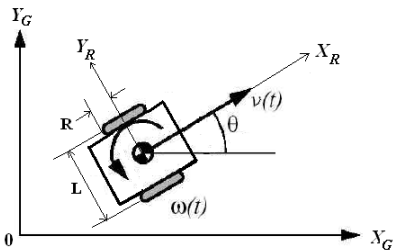

Figure 7. The robot's pose and parameters

We start with the kinematic model of the mobile robot. The two wheeled, differential-drive mobile robot with non-slipping and pure rolling is considered. Fig.7 shows the coordinate systems and notations for the robot, where ($X_G$, $Y_G$) is the global coordinate, ($X_R$, $Y_R$) is the local coordinate relative to the robot chassis. $R$ denotes the radius of driven wheels, and $L$ denotes the distance between the wheels.

During one sampling period $\Delta t$, the rotational speed of the left and right wheels $\omega_L$ and $\omega_R$ create corresponding increment distances $\Delta s_L$ and $\Delta s_R$ traveled by the left and right wheels respectively:

$$\Delta s_L = \Delta t R \omega_L \qquad \Delta s_R = \Delta t R \omega_R \qquad (1)$$

These can be translated to the linear incremental displacement of the robot's center $\Delta s$ and the robot's orientation angle $\Delta \theta$:

$$\Delta s = \frac{\Delta s_L + \Delta s_R}{2} \qquad \Delta \theta = \frac{\Delta s_R - \Delta s_L}{L} \qquad (2)$$

The coordinates of the robot at time $k+1$ in the global coordinate frame can be then updated by:

$$\begin{bmatrix} x_{k+1} \\ y_{k+1} \\ \theta_{k+1} \end{bmatrix} = \begin{bmatrix} x_k \\ y_k \\ \theta_k \end{bmatrix} + \begin{bmatrix} \Delta s_k \cos(\theta_k + \Delta \theta_k / 2) \\ \Delta s_k \sin(\theta_k + \Delta \theta_k / 2) \\ \Delta \theta_k \end{bmatrix} \qquad (3)$$

In practice, (3) is not really accurate due to unavoidable errors appeared in the system. Errors can be both systematic such as the imperfectness of robot model and nonsystematic such as the slip of wheels. These errors have accumulative characteristic so that they can break the stability of the system if appropriate compensation is not considered. In our system, the compensation is carried out by the EKF.

Let $\mathbf{x} = [x\, y\, \theta]^T$ be the state vector. This state can be observed by some absolute measurements, $\mathbf{z}$. These measurements are described by a nonlinear function, $h$, of the robot coordinates and an independent Gaussian noise process, $\mathbf{v}$. Denoting the function (3) is $f$, with an input vector $\mathbf{u}$, the robot can be described by:

$$\begin{aligned} \mathbf{x}_{k+1} &= f(\mathbf{x}_k, \mathbf{u}_k, \mathbf{w}_k) \\ \mathbf{z}_k &= h(\mathbf{x}_k, \mathbf{v}_k) \end{aligned} \qquad (4)$$

where the random variables $\mathbf{w}_k$ and $\mathbf{v}_k$ represent the process and measurement noise respectively. They are assumed to be independent to each other, white, and with normal probability distributions: $\mathbf{w}_k \sim N(0, \mathbf{Q}_k) \quad \mathbf{v}_k \sim N(0, \mathbf{R}_k) \quad E(\mathbf{w}_i \mathbf{v}_j^T) = 0$

The steps to calculate the EKF are then realized as below:
1. Prediction step with time update equations:

$$\hat{\mathbf{x}}_k^- = f(\hat{\mathbf{x}}_{k-1}, \mathbf{u}_{k-1}, \mathbf{0}) \qquad (5)$$

$$\mathbf{P}_k^- = \mathbf{A}_k \mathbf{P}_{k-1} \mathbf{A}_k^T + \mathbf{W}_k \mathbf{Q}_{k-1} \mathbf{W}_k^T \qquad (6)$$

where $\hat{\mathbf{x}}_k^- \in \mathfrak{R}^n$ is the *priori* state estimate at step $k$ given knowledge of the process prior to step $k$, $\hat{\mathbf{P}}_k^-$ denotes the covariance matrix of the state-prediction error, $\mathbf{A}_k$ is the Jacobian matrix of partial derivates of $f$ to $x$:

$$\mathbf{A}_{ij} = \frac{\partial \mathbf{f}_i}{\partial \mathbf{x}_{pj(k-1)}}|_{(\hat{\mathbf{x}}_{p(k-1)}, \mathbf{u}_{(k-1)}, \mathbf{0})}; \mathbf{A}_k = \begin{bmatrix} 1 & 0 & -\Delta s_k \sin(\theta_k + \Delta \theta_k / 2) \\ 0 & 1 & \Delta s_k \cos(\theta_k + \Delta \theta_k / 2) \\ 0 & 0 & 1 \end{bmatrix} \qquad (7)$$

$\mathbf{W}$ is the Jacobian matrix of partial derivates of $f$ to $w$:

$$W_{k+1} = \frac{\partial f_k}{\partial \mathbf{w}}\bigg|_{(\hat{\mathbf{x}}_k^+, \mathbf{u}_k, \mathbf{0})} = \Delta t \frac{R}{2} \begin{bmatrix} \cos\left(\theta_k + \frac{\Delta\theta_k}{2}\right) - \frac{1}{L}\Delta s_k \sin\left(\theta_k + \frac{\Delta\theta_k}{2}\right) & \cos\left(\theta_k + \frac{\Delta\theta_k}{2}\right) + \frac{1}{L}\Delta s_k \sin\left(\theta_k + \frac{\Delta\theta_k}{2}\right) \\ \sin\left(\theta_k + \frac{\Delta\theta_k}{2}\right) + \frac{1}{L}\Delta s_k \cos\left(\theta_k + \frac{\Delta\theta_k}{2}\right) & \sin\left(\theta_k + \frac{\Delta\theta_k}{2}\right) - \frac{1}{L}\Delta s_k \cos\left(\theta_k + \frac{\Delta\theta_k}{2}\right) \\ \frac{1}{L} & -\frac{1}{L} \end{bmatrix} \quad (8)$$

and $\mathbf{Q}_{k-1}$ is the input-noise covariance matrix which depends on the standard deviations of noise of the right-wheel rotational speed and the left-wheel rotational speed. They are modeled as being proportional to the rotational speed $\omega_{R,k}$ and $\omega_{L,k}$ of these wheels at step $k$. This results in the variances equal to $\delta\omega_R^2$ and $\delta\omega_L^2$, where $\delta$ is a constant determined by experiments. The input-noise covariance matrix $\mathbf{Q}_k$ is defined as:

$$\mathbf{Q}_k = \begin{bmatrix} \delta\omega_{R,k}^2 & 0 \\ 0 & \delta\omega_{L,k}^2 \end{bmatrix} \quad (9)$$

2. Correction step with measurement update equations:

$$\mathbf{K}_k = \mathbf{P}_k^- \mathbf{H}_k^T (\mathbf{H}_k \mathbf{P}_k^- \mathbf{H}_k^T + \mathbf{R}_k)^{-1} \quad (10)$$

$$\hat{\mathbf{x}}_k = \hat{\mathbf{x}}_k^- + \mathbf{K}_k \left( \mathbf{z}_k - \mathbf{h}(\hat{\mathbf{x}}_k^-) \right) \quad (11)$$

$$\mathbf{P}_k = (\mathbf{I} - \mathbf{K}_k \mathbf{H}_k) \mathbf{P}_k^- \quad (12)$$

where $\hat{\mathbf{x}}_k \in \Re^n$ is the *posteriori* state estimate at step $k$ given measurement $\mathbf{z}_k$, $\mathbf{K}_k$ is the Kalman gain, $\mathbf{H}$ is the Jacobian matrix of partial derivates of $h$ to $x$:

$$\mathbf{H}_{ij} = \frac{\partial \mathbf{u}_i}{\partial \hat{\mathbf{x}}_{pj(k-1)}}\big|_{(\hat{\mathbf{x}}_{p(k)})}; \mathbf{H}_k = \begin{bmatrix} -C_1\cos(\beta_1) & -C_1\sin(\beta_1) & 0 \\ 0 & 0 & -1 \\ . & . & . \\ . & . & . \\ . & . & . \\ -C_N\cos(\beta_N) & -C_N\sin(\beta_N) & 0 \\ 0 & 0 & -1 \\ 0 & 0 & 1 \end{bmatrix} \quad (13)$$

$\mathbf{R}_k$ is the covariance matrix of noises estimated from the measurements of compass sensor and LRF as follows.

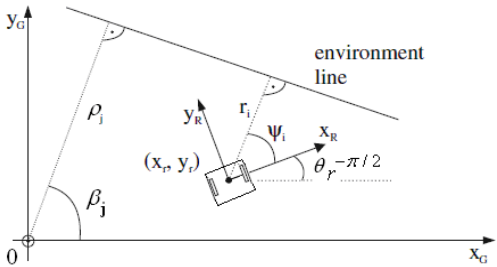

Figure 8. The line parameters ($\rho_j$, $\beta_j$) according to the global coordinates, and the line parameters ($r_i$, $\psi_i$) according to the robot coordinates

At the first scan of LRF, a global map of environment is constructed composed of a set of line segments described by parameters $\beta_j$ and $\rho_j$. The line equation in normal form is:

$$x_G \cos\beta_j + y_G \sin\beta_j = \rho_j \quad (14)$$

When the robot moves, a new scan of LRF is performed and a new map of environment, namely local map, is constructed which also consists of a set of line segments described by the equation:

$$x_R \cos\psi_i + y_R \sin\psi_i = r_i \quad (15)$$

where $\psi_i$ and $r_i$ are the parameters of lines (fig.8).

The line segments of the global and local map are then matched using *weighted line fitting algorithm* [11]. The matching line parameters $r_i$ and $\psi_i$ from the local map are collected in the vector $\mathbf{z}_k$, which is used as the input for the correction step of the EKF to update the robot's state:

$$\mathbf{z}_k = [r_1, \psi_1, ....., r_N, \psi_N, \varphi_k]^T \quad (16)$$

where $\varphi_i$ is the signal measured from the compass sensor.

From the robot pose estimated by odometry, the parameters $\rho_j$ and $\beta_j$ of the line segments in the global map (according to the global coordinates) is transformed into the parameters $\hat{r}_i$ and $\hat{\psi}_i$ (according to the coordinates of the robot) by:

$$C_j = \rho_j - x_{r(k)}\cos(\beta_j) - y_{r(k)}\sin(\beta_j) \quad (17)$$

$$\begin{bmatrix} \hat{r}_i \\ \hat{\psi}_i \\ \hat{\theta}_k \end{bmatrix} = \mathbf{u}(\hat{x}_k^-, \rho_j, \beta_j, \varphi_i) = \begin{bmatrix} |C_j| \\ \beta_j - (\hat{\theta}_i^- - \pi/2) + (-0.5\,\mathrm{sign}(C_j) + 0.5)\pi \\ \hat{\theta}_k^- \end{bmatrix} \quad (18)$$

The covariance matrix $\mathbf{R}_{i,k}$ of measurement noise has the diagonal structure, where the $i^{th}$ block is:

$$\mathbf{R}_{i,k} \cong \begin{bmatrix} \mathrm{var}(r_i) & 0 & ... & 0 & 0 \\ 0 & \mathrm{var}(\psi_i) & ... & ... & ... \\ ... & ... & ... & ... & ... \\ ... & ... & ... & \mathrm{var}(\psi_i) & 0 \\ 0 & 0 & ... & 0 & \mathrm{var}(\varphi_k) \end{bmatrix} \quad (19)$$

From (16) and (19), the data of the compass sensor is utilized to correct the robot orientation. At step $k$, this data is employed as the absolute measurement for the element $\theta_k$ of $\mathbf{z}_k$. The noise of this measurement is achieved from the manufactory specification. The accuracy of $0.1^0$ corresponds to the noise variance of $0.01$. This collected into the covariance matrix $\mathbf{R}_{i,k}$ (19) gives $\mathbf{R}_k$ for the correction step of EKF.

## IV. EXPERIMENTS

The setup of the sensor system has been described in Section II. In this section, experiments are conducted to evaluate the efficiency of the fusion algorithm.

## A. Experimental Setup

A rectangular shaped flat-wall environment constructed from several wooden plates surrounded by a cement wall is setup. The robot is a two wheeled, differential-drive mobile robot. Its wheel diameter is *10 cm* and the distance between two drive wheels is *60cm*.

The speed stability of the motor during each sampling time step is an important factor to maintain the efficiency of the EKF. For that reason, motors are controlled by microprocessor-based electronic circuits which carry out a PID algorithm. The stability checked by a measuring program written in LABVIEW is *±5%*.

The compass sensor has the accuracy of $0.1^0$. The LRF has the accuracy of *30mm* in distance and $0.25^0$ in deflect angle. The sampling time $\Delta t$ of the EKF is *100ms*. The proportional factor $\delta$ of the input-noise covariance matrix $\mathbf{Q}_k$ is experimentally estimated as follows. The deviation between the true and the position estimated by the kinematic model when driving the robot straight forwards several times (from the minimum to the maximum tangential speed of the robot) is determined. The deviation between the true robot orientation and the orientation estimated by the kinematic model when rotating the robot around its own axis several times (from the minimum to the maximum angular speed of the robot) is also determined. Based on the error in position and orientation, the parameter $\delta$ is calculated. In our system, $\delta$ is estimated to be the value 0.01.

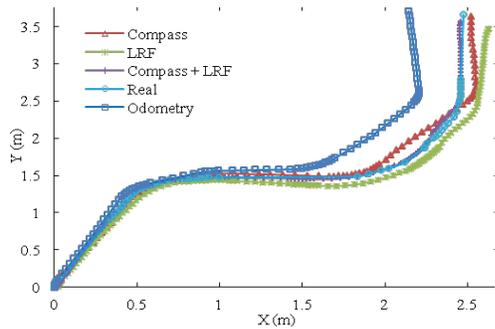

Figure 9. Estimated robot trajectories under different EKF configurations

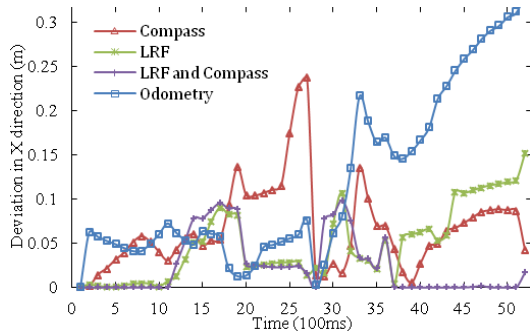

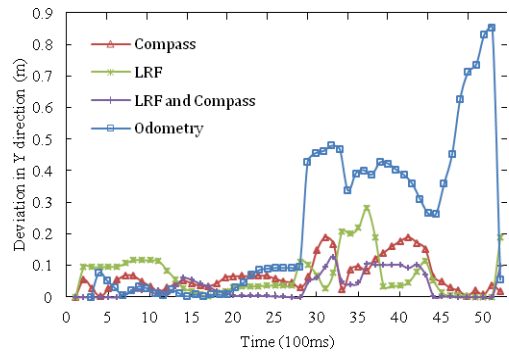

Figure 10. The deviation in X and Y direction between estimated positions and the real one

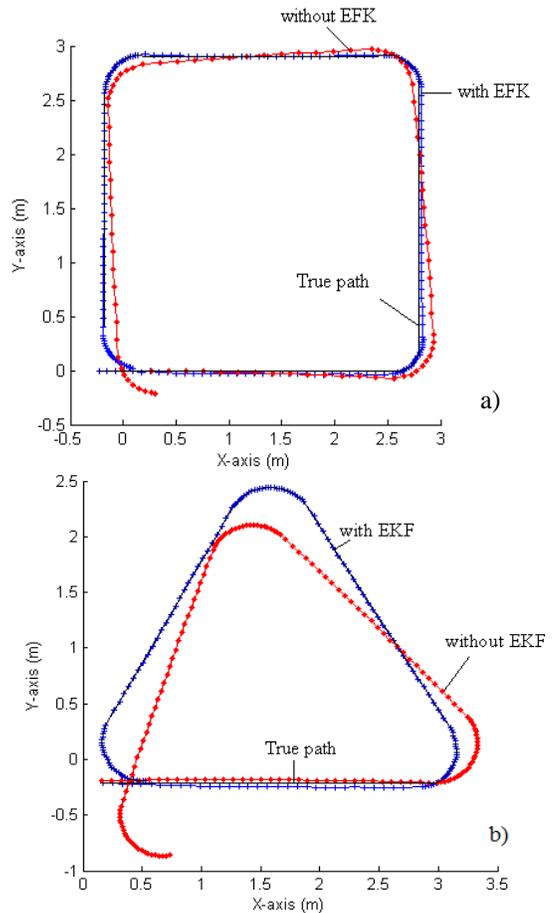

Figure 11. Trajectories of the robot moving along predefined paths
a) Rounded rectangular path   b) Rounded triangular path

## B. Sensor Fusion Algorithm Evaluation

In order to evaluate the efficiency of the fusion algorithm, different configurations of the EKF were implemented. Fig.9 describes the trajectories of a robot movement estimated by four methods: the odometry, the EKF with compass sensor, the EKF with LRF, and the EKF with the combination of LRF and compass sensor. The deviations between each trajectory and the real one are represented in fig.10.

In another experiment, the robot is programmed to follow predefined paths under two different scenarios: with and without the EKF. Fig.11a represents the trajectories of the robot moving along a rounded rectangular path in which the one with dots corresponds to the non-existence of EKF and the one with pluses corresponds to the existence of EKF in the implementation. The trajectories in case the robot follows a rounded triangular path are shown in fig.11b.

It is concluded that the EKF algorithm improves the robot localization and its combination with LRF and compass sensor gives the optimal result.

*C. Autonomous Navigation*

In this experiment, we evaluate the applicability of the proposed fusion approach in a real autonomous navigation application. The goal is to navigate the mobile robot to go a round closed to the line segments of the global map extracted from the first scan of the LRF. Fig.12 shows the extracted map and the success trajectory of the robot during the navigation. A sequence of images showing the motion of the robot in the experiment is described in fig.13.

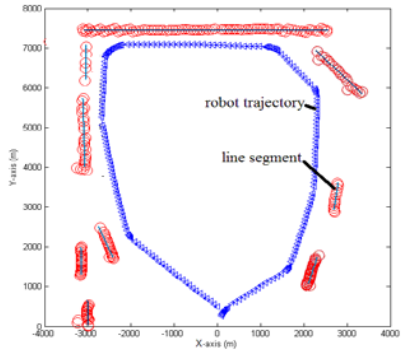

Figure 12. Line segments of a global map and the trajectory of the robot

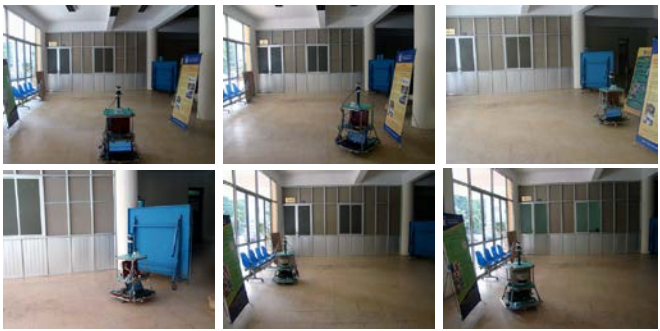

Figure 13. A sequence of images showing the motion of robot in an experimental environment during the autonomous navigation operation

V. CONCLUSION

It is necessary to develop a perceptual system for the mobile robot. The system is required to not only be well-working but also critically support various levels of perception. In this work, many types of sensors including position speed encoders, integrated sonar ranging sensors, compass and laser finder sensors, the global positioning system (GPS) and the visual system have been implemented in a real mobile robot platform. An EKF has been designed to fuse the raw data of compass sensor and LRF. Experiments show that this novel combination significantly improves the accuracy of robot localization and is sufficient to ensure the success of robot navigation. Further investigation will be continued with more sensor combination to better support the localization in outdoor environments.

ACKNOWLEDGMENT

This work was supported by Vietnam National Foundation for Science and Technology Development (NAFOSTED).